\documentclass[letterpaper, 10 pt, conference]{ieeeconf}  

\IEEEoverridecommandlockouts                              

\usepackage{cite}
\usepackage{caption}
\usepackage{subcaption}
\usepackage{amsmath,amssymb,amsfonts}
\usepackage[ruled, vlined]{algorithm2e}
\usepackage{lipsum}
\usepackage{graphicx}
\usepackage{textcomp}
\usepackage{xcolor}
\usepackage{balance}
\usepackage{pgfplots}
\usepackage[per-mode=fraction]{siunitx}
\DeclareSIUnit\pixel{px}
\usepackage{comment}
\usepackage{multirow}
\usepackage{hhline}
\usepackage{svg}
\usepackage{diagbox}

\makeatletter
\let\NAT@parse\undefined
\makeatother
\usepackage{hyperref}  
\usepackage{cleveref}

\usepackage[absolute]{textpos}
\newcommand{\copyrightstatement}{
	\begin{textblock*}{17cm}(20mm,1mm)    
		\noindent
		\footnotesize
		\copyright 2024 IEEE. Personal use of this material is permitted. Permission from IEEE must be
		obtained for all other uses, in any current or future media, including
		reprinting/republishing this material for advertising or promotional purposes, creating new
		collective works, for resale or redistribution to servers or lists, or reuse of any copyrighted
		component of this work in other works. DOI: 10.1109/IV55156.2024.10588831
	\end{textblock*}
}

\DeclareMathOperator{\atantwo}{atan2}

\DeclareUnicodeCharacter{2212}{−}
\usepgfplotslibrary{groupplots,dateplot}
\usetikzlibrary{patterns,shapes.arrows}
\pgfplotsset{compat=newest}

\title{\LARGE \bf
Low Latency Instance Segmentation by Continuous Clustering for LiDAR Sensors
}

\author{Andreas Reich and Mirko Maehlisch
\thanks{The authors gratefully acknowledge funding by the Federal Office of Bundeswehr Equipment, Information Technology and In-Service Support (BAAINBw).}
\thanks{Both authors are with Institute for Autonomous Systems Technology (TAS),
        University of the Bundeswehr Munich, Neubiberg, Germany
        {\tt\small andreas.reich@unibw.de}}%
}

\begin{document}
\copyrightstatement
\bstctlcite{bibcontrol_etal4}

\maketitle

\thispagestyle{empty}
\pagestyle{empty}

\begin{abstract}


Low-latency instance segmentation of LiDAR point clouds is crucial in real-world applications because it serves as an initial and frequently-used building block in a robot's perception pipeline, where every task adds further delay.
Particularly in dynamic environments, this total delay can result in significant positional offsets of dynamic objects, as seen in highway scenarios.
To address this issue, we employ a new technique, which we call continuous clustering.
Unlike most existing clustering approaches, which use a full revolution of the LiDAR sensor, we process the data stream in a continuous and seamless fashion.
Our approach does not rely on the concept of complete or partial sensor rotations with multiple discrete range images; instead, it views the range image as a single and infinitely horizontally growing entity.
Each new column of this continuous range image is processed as soon it is available.
Obstacle points are clustered to existing instances in real-time and it is checked at a high-frequency which instances are completed in order to publish them without waiting for the completion of the revolution or some other integration period.
In the case of rotating sensors, no problematic discontinuities between the points of the end and the start of a scan are observed.
In this work we describe the two-layered data structure and the corresponding algorithm for continuous clustering.
It is able to achieve an average latency of just 5 ms with respect to the latest timestamp of all points in the cluster.
We are publishing the source code at \url{https://github.com/UniBwTAS/continuous_clustering}.

\end{abstract}

\section{INTRODUCTION}

\begin{figure}[t]
     \centering
     \begin{subfigure}[b]{\linewidth}
        \centering
        \includegraphics[width=\linewidth]{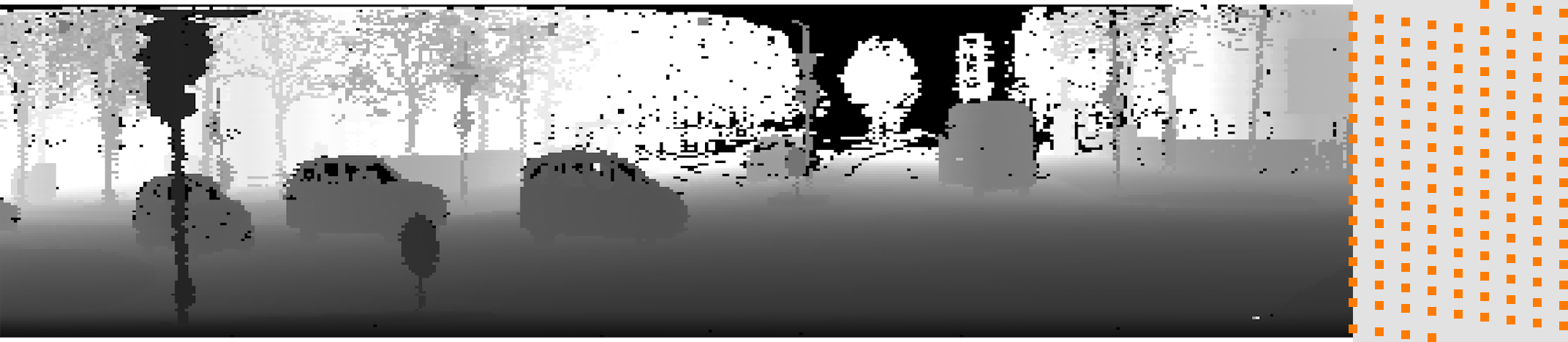}
        \caption{Extract of infinite range image with single firing (orange)}
        \label{fig:title_a}
     \end{subfigure}
     \hfill
     \begin{subfigure}[b]{\linewidth}
         \centering
         \includegraphics[trim={0 0 0 80}, clip, width=\linewidth]{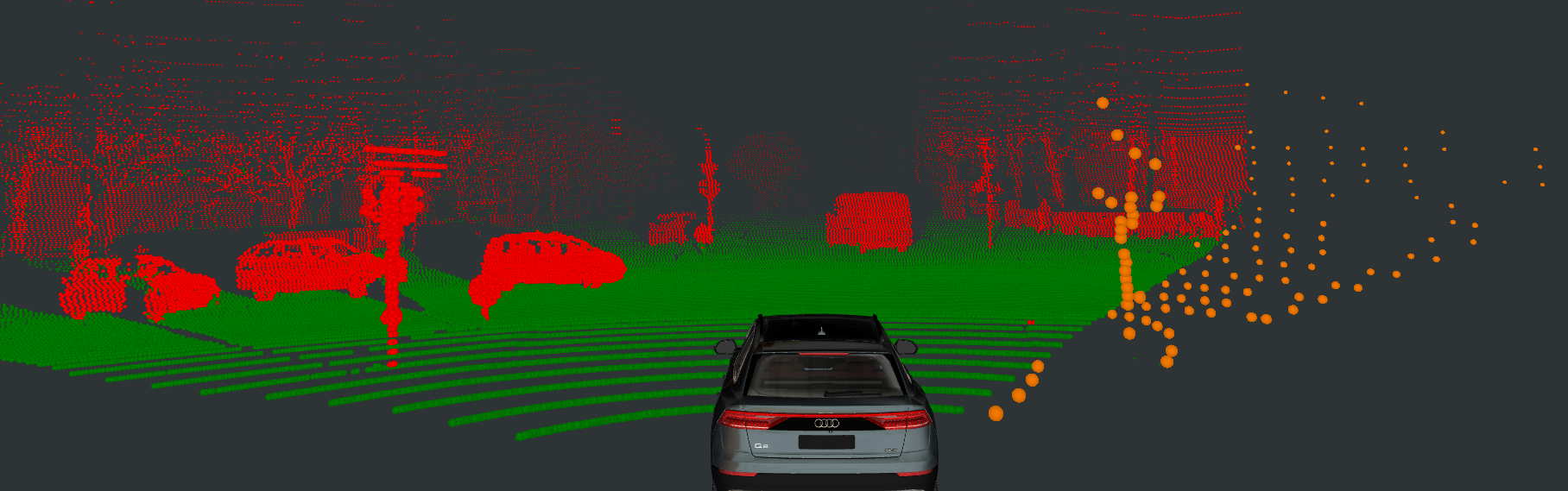}
         \caption{Online ground point classification with single firing (orange)}
         \label{fig:title_b}
     \end{subfigure}
     \begin{subfigure}[b]{\linewidth}
         \centering
         \includegraphics[trim={0 0 0 80}, clip, width=\linewidth]{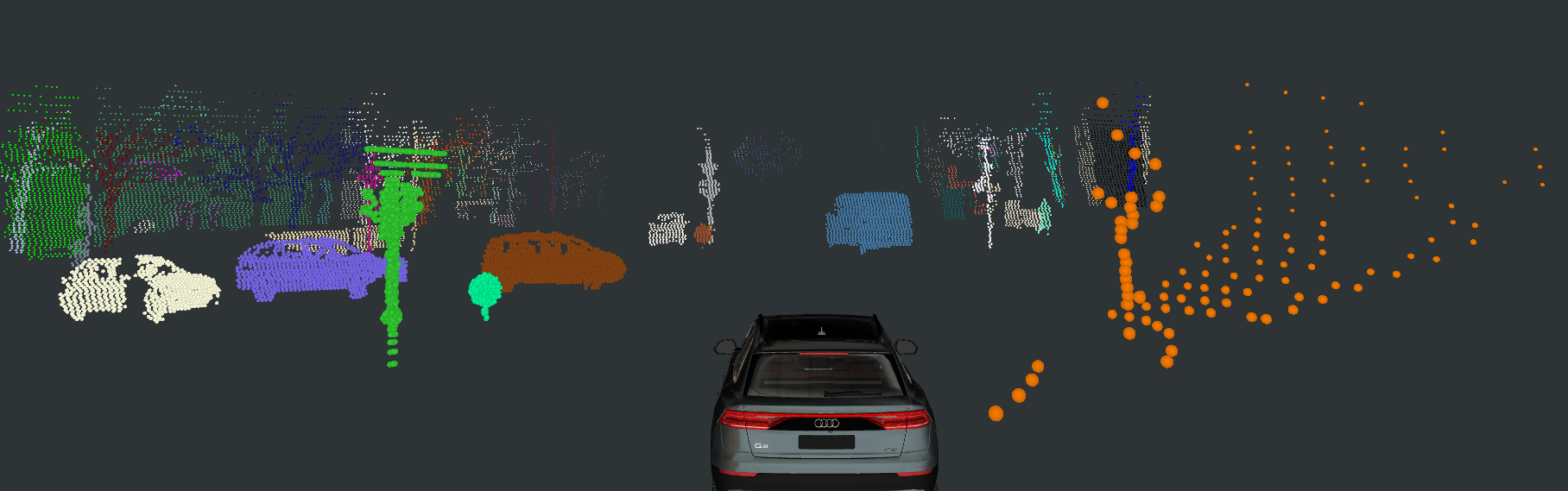}
         \caption{Clustering output (published as soon as all lasers have passed)}
         \label{fig:title_c}
     \end{subfigure}
    \caption{
\textbf{Continuous clustering:}
(a) An infinite range image is constructed from a continuous stream of firings (a single shot from all laser diodes).
Each point within a firing has a specific constant azimuth offset in order to reduce interference between laser beams.
(b) Once a column in the range image is complete (rearmost laser of firing has passed the column), the points are classified as ground (green) and obstacle points (red).
(c) The obstacle points are clustered if their mutual distance is below a threshold $d_\text{T}$ and published as soon as no more points can be added (rearmost laser is distant enough).
}
    \label{fig:title}
\end{figure}

Instance segmentation for LiDAR points is crucial in robotics, as it enables the precise recognition and differentiation of objects in 3D space, facilitating safer and more efficient navigation and interaction with the environment.
The resulting instances can be processed in downstream perception tasks, such as multi-object tracking based on the tracking-by-detection framework or landmark-based simultaneous localization and mapping (SLAM).

A simple but effective and verifiable approach for instance segmentation is Euclidean distance-based clustering of LiDAR points.
Thus, any two points are clustered if their mutual distance is below a threshold $d_\text{T}$.
Another major advantage is that it can detect any type of object class unlike typical supervised learning-based instance segmentation approaches, which are trained to recognize a predefined set of object classes.

We extend this concept to continuous clustering as visualized in \Cref{fig:title}.
This means that the algorithm does not only start after a full rotation of the LiDAR sensor, but directly starts extracting individual instances as soon as they are complete.
Our approach does not rely on the concept of complete or partial sensor rotations with multiple discrete range images; instead, it views the range image as a single and infinitely horizontally expanding unit.
Thus, our method is not limited to LiDAR sensors that rotate around a single axis but can also be applied to LiDAR sensors that rotate around multiple axes \cite{karimi_2021_lola}.
With slight modifications, this method is also applicable to horizontally scanning solid-state LiDAR sensors in order to reduce latency.

To the best of our knowledge, there is currently no clustering approach capable of identifying and publishing finished instances from a continuous stream of points before the rotation or some other integration period of the LiDAR sensor is completed.
We further elaborate on this claim in \Cref{sec:related_work}.
Consequently, this may be regarded as our principal contribution.
In addition, despite achieving sufficient throughput on consumer grade hardware, we present two additional heuristics aimed at further accelerating the clustering algorithm to enable real-time processing on hardware with limited computation power.
However, it should be noted that under these conditions, the guarantee of an exact result is no longer feasible. We provide a short video\footnote{https://www.mucar3.de/iv2024-continuous-clustering} illustrating the capability of our method.

\section{RELATED WORK}
\label{sec:related_work}

We begin by describing approaches that divide the LiDAR rotation into slices to reduce latency.
Subsequently, we detail existing clustering-based methods that operate on a full LiDAR rotation, to convey the foundational concepts.
Lastly, we introduce two further approaches that most closely resemble our work in that they initiate clustering prior to the completion of a LiDAR rotation and are based on clustering.

\paragraph{Slice-based Approaches for Latency Reduction}
There are several machine learning techniques capable of processing slices of a rotation \cite{han_2020_streaming, chen_2021_polarstream, loiseau_2022_online, abdelfattah_2023_multi}.
There is a trade-off between reducing latency through smaller slices and minimizing cuts within object instances by enlarging the slices.
In order to handle these cuts, spatio-temporal network architectures are utilized, such as recurrent neural networks, transformer-based architectures or temporally accumulated bird's-eye view feature maps.
Since our approach is not based on discrete segments, this issue does not arise here.
Although our work discusses clustering based solely on Euclidean distance, it is also conceivable to use the similarity of a point's feature vector as additional criterion, where distance serves merely as an upper threshold.
This enables our approach to be applicable in the context of deep learning methodologies.

\paragraph{Clustering-based Approaches on Full Scan}
Our work builds upon the concepts of existing distance-based clustering approaches on range images \cite{moosmann_2009_segmentation, bogoslavskyi_2016_fast, zermas_2017_fast, burger_2018_fast2, yang_2020_two, zhang_2022_real, oh_2022_travel}.
All of them first accumulate a full LiDAR rotation before they start to generate a range image.
Almost all methods use the metadata and the ordered data stream to sort the points into a range image, where rows correspond to individual lasers sorted by elevation angle, and columns contain points within a particular azimuth range \cite{moosmann_2009_segmentation, bogoslavskyi_2016_fast, zermas_2017_fast, burger_2018_fast2, yang_2020_two, zhang_2022_real}.
Compared to spherical projection based methods \cite{oh_2022_travel}, this leads to a complete and, at the same time, compact data representation with a low number of empty cells and insertion collisions.
Therefore, in our approach the continuous range image is also generated by extending the concepts of this method.

In the next step, most approaches carry out a ground point segmentation \cite{bogoslavskyi_2016_fast, zermas_2017_fast, burger_2018_fast2, yang_2020_two, zhang_2022_real, oh_2022_travel}.
This is especially crucial for purely distance-based clustering; otherwise, all objects would be merged across the ground plane.
Only \cite{moosmann_2009_segmentation} forgoes this segmentation, treating the ground points as a separate cluster.
This, however, requires an adjusted metric in the subsequent stage to prevent generating edges between ground and obstacle points in the neighborhood graph.
Our approach also removes the ground points in order to reduce the work load and the complexity in subsequent steps.

In the next step, edges are inserted or deleted between the neighborhood graph, whose vertices correspond to the obstacle points obtained in the previous step.
The range image is used to identify potential neighbors efficiently.
There are purely Euclidean distance-based approaches \cite{zermas_2017_fast, zhang_2022_real, oh_2022_travel} and those that introduce additional conditions \cite{moosmann_2009_segmentation, bogoslavskyi_2016_fast, burger_2018_fast2, yang_2020_two}.
As an additional condition, convexity is sometimes taken into account in order to separate two objects even if they are spatially close to each other \cite{moosmann_2009_segmentation, bogoslavskyi_2016_fast, yang_2020_two}.
In contrast, \cite{burger_2018_fast2} reintroduces edges between two adjacent cells of the range image when the points are actually too far apart but lie on a straight line.
This may occur, for example, with distant walls.
In our work, we consider only the Euclidean distance to keep the explanations as simple as possible. However, our approach can easily be extended to incorporate these concepts.

In the final step, all approaches commonly aim to identify the resulting connected components, which are subgraphs that aren't part of a larger graph. 
This is a well-known problem in graph theory and is typically referred to as connected component labeling (CCL).

\begin{figure*}
    \centering
    \includegraphics[width=\textwidth]{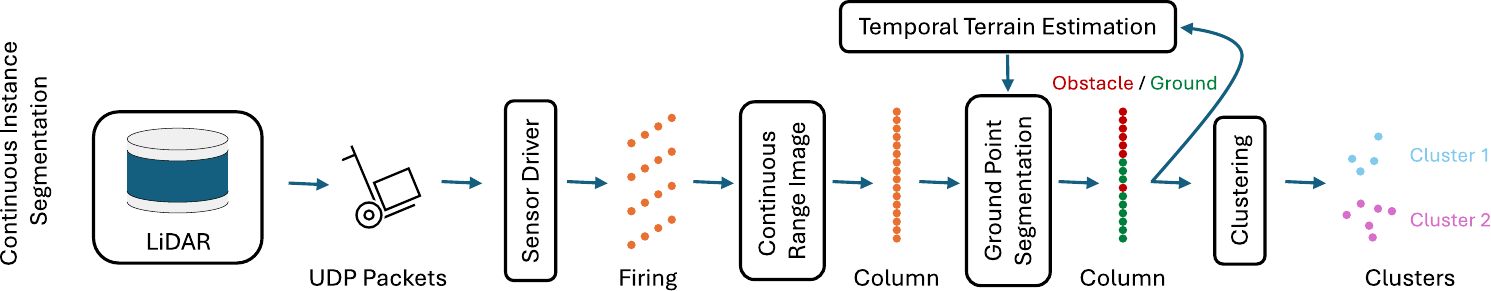}
    \caption{
    \textbf{Continuous Clustering Pipeline}
    }
    \label{fig:architecture}
\end{figure*}

\paragraph{Continuous Clustering-based Approaches}
To our knowledge, there are only two works \cite{najdataei_2018_continuous} and \cite{najdataei_2022_pi_lisco}, both from by Najdataei et al., that initiate clustering before a sensor revolution is completed.
To delineate our contributions, we elaborate on these works below and identify the distinctions.

In their initial study \cite{najdataei_2018_continuous}, the concept also aims at minimizing latency by initiating clustering before the completion of a LiDAR rotation.
Similarly, with each incoming column of the range image, points are clustered into existing instances or new instances are initiated as needed.
The data structure they utilized is akin to the lower layer of our two-layered data structure.
However, this approach does not explicitly identify complete clusters but instead waits for a full rotation to output all existing clusters.
Thus, it merely minimizes the time between the end of the rotation and the output of clusters, meaning the worst-case latency still equals the full rotation period $T$ if the relevant object is at the start of the rotation.
Our method is capable of identifying and outputting complete clusters as rapidly as possible during the rotation, as described in \Cref{sec:cluster_generation}.
Furthermore, the junction between the end and the beginning of a LiDAR rotation remains an issue, as it clusters in a continuous manner within a range image for a single rotation but not across range image boundaries.
Moreover, Najdataei et al. do not elaborate on how they handle LiDAR sensors whose laser diodes are not vertically aligned, common among many sensors.
They also do not discuss how ground point segmentation is performed with knowledge of only the current and past columns of a range image.
We detail these processes in \Cref{sec:continuous_range_image_generation} and \Cref{sec:ground_point_classification}, respectively.

In a subsequent work \cite{najdataei_2022_pi_lisco}, Najdataei et al. propose a method that is capable of reporting clustering results after every fraction $T_\text{frac}$ of a LiDAR rotation.
For example, clustering for the previous complete rotation can be output after every half $T_\text{frac} = 0.5T$ or quarter rotation $T_\text{frac} = 0.25T$.
Thus, they pursue a global continuous approach, eliminating junction issues.
The foundational concept is that the range images, depending on the reporting period $T_\text{frac}$, have a large intersection area, so ideally, the neighborhood graph only needs to be updated at the horizontal ends.
In our view, the main issue with this approach is that it outputs clustering for the preceding rotation each time, so depending on the reporting period $T_\text{frac}$, i.e. the intersection between the two range images, clusters may be output multiple times.
Especially if this approach is pushed to the limit and the reporting period $T_\text{frac}$ is continuously shortened, bandwidth to subsequent perception modules quickly becomes a bottleneck.
In our approach, the evaluation period $T_\text{frac}$ is only the time for one column, potentially delivering results at a peak frequency of $\SI{17000}{\hertz}$ for our sensor.
This is calculated from $10$ rotations per second and approximately $1700$ columns per rotation.
At this high frequency it is only feasible to output each cluster exactly once, ideally at the moment it is complete, i.e. the pairwise distance of any future point to any point in the cluster must be above $d_T$.
This also benefits subsequent perception modules by indicating whether a cluster is complete.
In the approach by Najdataei et al., the reporting also contains incomplete clusters.
Moreover, we argue that aside from the bandwidth issue, their clustering algorithm can only feasibly be executed at this frequency with very high parallelization.
This is particularly true for horizontally extended objects like walls, which result in very large so-called extended neighborhoods in the graph and thus many affected points.
With our data structure, new points can be inserted much more efficiently with significantly fewer updates in the existing neighborhood graph.
In their work it is described that ground point segmentation is performed based on a constant threshold for the z-coordinate of a point.
However, this is only applicable if the sensor is parallel to the ground surface and if the terrain has no inclines.

\section{PROPOSED METHOD}
\label{sec:proposed_method}

In the following, each stage of our continuous clustering pipeline is described.
This pipeline is visualized in \Cref{fig:architecture}.
The individual stages are designed to work in a pipelined fashion.
This means that the stages can be parallelized, i.e. processed by an individual thread, while passing the job from one stage to the next.
A horizontally continuous range image serves as the shared data structure.
As the memory is limited, it is implemented as a cyclic buffer, which stores the last $W$ columns in a queue and still allows random access to the individual cells.
The height of the range image corresponds to the number of lasers of the LiDAR sensor: $H=128$.

The first stage, described in \Cref{sec:continuous_range_image_generation}, inserts a firing into the continuous range image, visualized in \Cref{fig:title_a}.
It also identifies potentially completed columns, which are passed on to the ground point segmentation stage, explained in \Cref{sec:ground_point_classification}.
Here, each point is classified as obstacle or ground point.
In the next stage, these obstacle points of a column are aggregated and associated to point trees as described in \Cref{sec:point_tree_construction}.
These point trees represent disjoint subsets of the full cluster and are linked to each other if any of their points are closer than $d_T$.
In the final stage, explained in \Cref{sec:cluster_generation}, finished clusters are identified by analyzing all their unpublished point trees.

\subsection{Continuous Range Image Generation}
\label{sec:continuous_range_image_generation}

A 2D range image drastically reduces the computational complexity in order to find neighbor points compared to an unordered point cloud.
We use the concepts of \cite{moosmann_2009_segmentation, bogoslavskyi_2016_fast, zermas_2017_fast, burger_2018_fast2, yang_2020_two, zhang_2022_real, reich_2022_fast} and extend them to the continuous case.

The row index $i_\text{row}$ of a point is equal to the laser diode index ordered by their elevation angle.
Points generated by the laser diode with the highest elevation angle are inserted at the top row of the range image.

In our case the sensor is rotating clockwise, which means that new points are added to the right of the existing range image.
A point's column index $i_\text{col}$ is calculated based on its continuous azimuth angle $\varphi_\text{cont}$.
Compared to a points regular azimuth angle $\varphi=\atantwo(y, x)$, the value of $\varphi_\text{cont}$ monotonically increases with ongoing rotations of the LiDAR sensor. It can be calculated as $\varphi_\text{cont} = (-\varphi + \pi) + 2 \pi i_\text{rot}$, where $i_\text{rot}$ is the rotation index and $(x, y, z)$ is the Cartesian position of the current point in the sensor's coordinate system.
This ensures, that $\varphi_\text{cont}$ is monotonically increasing for a laser diode in a clockwise rotating sensor despite the opposite direction and the cyclic nature of $\atantwo$.
This simplifies calculations in later processing steps.

The global column index $i_\text{col}$ of a point in the continuous range image is also monotonically and indefinitely increasing.
It is defined as $i_\text{col} = \lfloor \varphi_\text{cont} / \Delta\varphi_\text{col} \rfloor$, where $\Delta\varphi_\text{col} = 2\pi / N_\text{firings}$ is the angular column width and $N_\text{firings}$ is a constant representing approximately the predicted number of firings per rotation.
Our sensor approximately generates $1800$ firings per rotation: $N_\text{firings} = 1800$.
Since the global column index is always increasing, it has to be mapped to the local column index of the cyclic buffer storing the continuous range image.
This step, however, is omitted in the following to keep the explanation simple.

Finally, all columns whose column index $i_\text{col} < i_\text{col, rear}$, where $i_\text{col, rear} = \lfloor \varphi_\text{cont, rear} / \Delta\varphi_\text{col} \rfloor$, are considered to be finished. Here, $i_\text{col, rear}$ and $\varphi_\text{cont, rear}$ are the global column index and the continuous azimuth angle of the latest firing's rearmost laser with respect to the rotation direction, respectively.

\subsection{Ground Point Classification}
\label{sec:ground_point_classification}

\begin{figure}
     \centering
     \begin{subfigure}[b]{\linewidth}
        \centering
        \includegraphics[width=\linewidth]{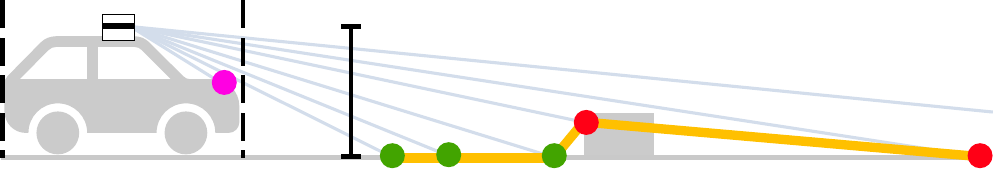}
        \caption{Ground points found in column (dark green)}
        \label{fig:ground_first_pass}
     \end{subfigure}
     \hfill
     \begin{subfigure}[b]{\linewidth}
         \centering
         \includegraphics[width=\linewidth]{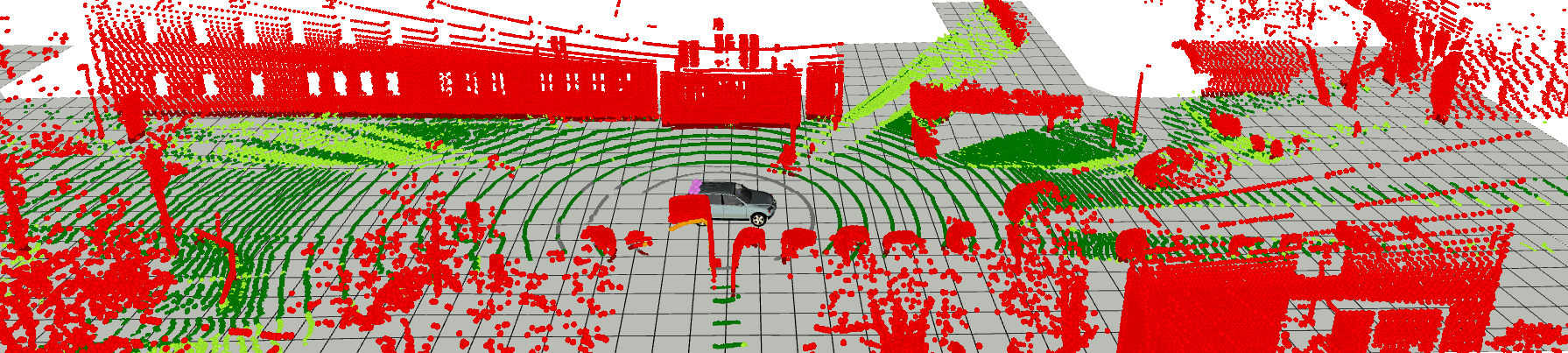}
         \caption{Further ground points (light green) with terrain}
         \label{fig:ground_second_pass}
     \end{subfigure}
    \caption{
    \textbf{
    Ground Point Classification:} 
    In the first pass (a) quite certain ground points are labeled based on the ego vehicle's bounding box, the sensor's height over ground and the slope between consecutive points in a column.
    In the second pass (b) remaining points are as labeled as ground, if their absolute z distance to the estimated terrain (gray) is below a threshold. The terrain estimation \cite{forkel_2021_probabilistic} gets the certain ground points (a) of the previous time steps as input.
    }
    \label{fig:ground_point_explanation}
\end{figure}

For ground point classification we use similar concepts to \cite{reich_2022_fast}.
A column is processed from bottom to top as visualized in \Cref{fig:ground_first_pass}.
Points inside the ego vehicle's bounding are omitted (magenta points).
If the $z$ coordinate of the first outside point matches the minimum of the ego robot, the point is classified as ground.
Subsequent points are also labeled as ground if the previous point was a ground point and the slope (yellow line) is small enough.
When calculating the slope it is beneficial to use the point's position in a world-fixed frame whose $xy$-plane is approximately aligned with the ground plane.
In this frame a slope near zero still means a flat line even if the sensor is mounted with a downward pitch angle.

Since not all ground points are found with this strategy, the algorithm in \cite{reich_2022_fast} also searches for ground points in the rows of the range image.
This is more challenging in our case as only the current and previous columns are known at this point in time.
Therefore, we use a terrain estimation module as described in \cite{forkel_2021_probabilistic}, which uses the certain ground points from the first pass for input.
It accumulates these ground points over time and generates a world-fixed estimation of the terrain.
This terrain is used to find remaining ground points based on the point's $z$ distance to this terrain.
The terrain is shown in \Cref{fig:ground_second_pass}.

\subsection{Point Tree Construction}
\label{sec:point_tree_construction}

\begin{figure*}
    \centering
    \includegraphics[width=\textwidth]{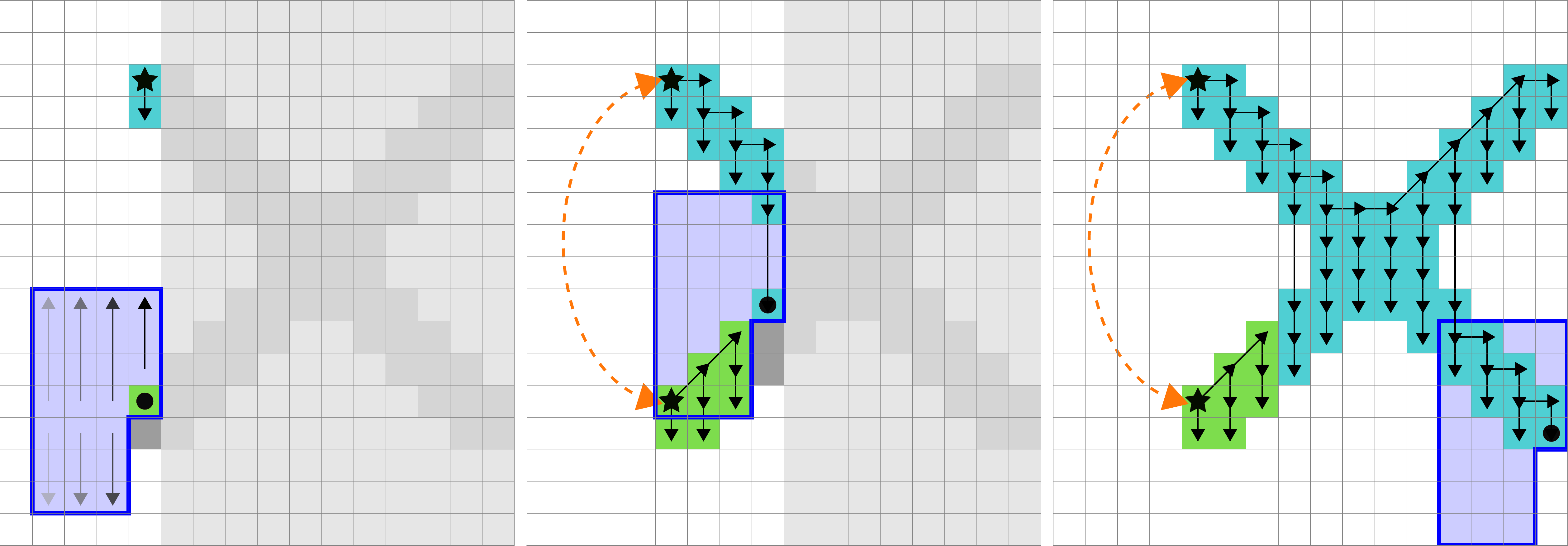}
    \caption{
    \textbf{Clustering of a X-shaped object:} Each incoming column of the range image is processed from top to bottom.
    The required size of the field of view (FOV, blue area) for the current point $p$ (black circle) is calculated such that it includes all neighbor points $p'$, which potentially could have a distance smaller than the distance threshold: $\lVert \overrightarrow{pp'} \rVert < d_\text{T}$.
    More specifically, all cells have to be evaluated whose angular difference to the current cell is smaller than $\arcsin(d_\text{T} / r_p)$, where $r_p$ is the current point's radius w.r.t. sensor origin.
    The cells inside the FOV are evaluated as indicated by the arrows (from dark to bright).
    If no neighbor is found, the current point becomes the root of a new point tree (black star).
    Otherwise, it becomes the child (black arrow) of the neighbor found first.
    For the point in the second image, the FOV includes neighbors from different trees with distance below $d_\text{T}$.
    This means, that the point trees belong to the same cluster and therefore a mutual higher-level link (orange) is generated.
    This results in a high-level graph, where the vertices are point trees.
    }
    \label{fig:graph_of_point_trees}
\end{figure*}

The main challenge for continuous clustering is the requirement to process the data in real-time.
This holds for both, the regular clustering as well as the additional challenge to evaluate at a high frequency which clusters are completed and therefore are ready to be published.
Therefore, we use a novel two-layered hierarchical data representation for this task.
The bottom layer are point trees $P_v$ and the top layer is an undirected and cyclic graph $G = (V, E)$, whose vertices $v \in V$ correspond to the point trees $P_v$ of the bottom layer.

\subsubsection{Point Trees}
\label{sec:point_tree}
A point tree $P_v$ grows from top to bottom and from left to right.
They are visualized by small black arrows in \Cref{fig:graph_of_point_trees}.
In this synthetic example there are two point trees (cyan and green), whose root cells are marked by a black star.
When a new column is processed, the algorithm iterates through the cells $p$ from top to bottom and searches for potential neighbors $p'$ in the point's field of view (blue area).
If a neighbor cell $p'$ with $\lVert \overrightarrow{pp'} \rVert < d_\text{T}$ was found, then $p$ is added to $p'$ as its child.
For this, each point $p$ maintains a list of 2D range image indices $p_\text{children}$ pointing to its child cells.
If no point was found the point becomes the root of a new point tree (green cell in the first step of \Cref{fig:graph_of_point_trees}).
This data structure is similar to the one presented in the existing approach from Najdataei et al. in \cite{najdataei_2018_continuous}.

\subsubsection{High-Level Graph Connecting Point Trees}
\label{sec:high_level_graph}
The high-level graph $G$ is only required to efficiently handle the edge case where two initially separated point trees grow together as more columns are revealed.
This case is visualized in step two of \Cref{fig:graph_of_point_trees}.
In the single layered approach by Najdataei et al. in \cite{najdataei_2018_continuous} the points of one point tree are relabeled such that all points belong to the same root point.
Depending on the size of the point tree this can be quite inefficient.
We do not merge them but just add a new edge $e$ to the high-level graph $G$ representing the fact that both point trees belong to the same cluster.
This is visualized by the orange arrow in \Cref{fig:graph_of_point_trees}.
Future points $p$ are still added to to just one of those point trees, namely the one of the first neighbor cell $p'$ in the FOV.
When such a neighbor cell $p'$ was found, it is important to traverse the rest of the FOV in order to search for additional neighbor cells with different point trees.
If this scenario occurs, a new edge $e$ is added to the high-level graph $G$.
In step two of \Cref{fig:graph_of_point_trees} the first neighbor $p'$ belongs to the cyan point tree.
Later, more neighbor cells are found that belong to the green point tree, which triggers the addition of edge $e$ to graph $G$.

The data structure described so far does not allow to efficiently obtain the point tree of a neighbor cell $p'$ (cyan or green color in \Cref{fig:graph_of_point_trees}) or equivalently its tree root index.
By storing just the child cells $p_\text{children}$, as described in \Cref{sec:point_tree}, it is just possible to traverse the point trees in a top-down fashion.
Therefore, we additionally store for each point also its tree root index $p_\text{root}$ at the moment it is added to a point tree via its neighbor $p'$: $p_\text{root} = p'_\text{root}$.
Once a point's tree root index $p_\text{root}$ was set, it remains constant for the future.
It allows to access the tree root in a constant amount of time $\mathcal{O}(1)$ irrespective of a point tree's height.

In the following we provide additional details how $G$ is efficiently stored.
All vertices $v \in V$ of graph $G$ are stored in a linked list.
This linked list just contains the 2D indices to the unpublished point tree roots.
All remaining meta data is stored at the root cells in the range image.
This includes the set $E_v \subseteq E$, which represent the outgoing high-level edges of $v$.
They are updated whenever two or more different point trees in $p$'s FOV are found.

Furthermore, this meta data includes $\varphi_{\text{finished}, v}$, which represents the specific continuous azimuth angle at which a point tree can be considered completed as no more points can be added due to the angular distance.
It is updated whenever a point $p$ is added to the point tree and is calculated as $\varphi_{\text{finished}, v} \gets \max(\varphi_{\text{finished}, v}, \varphi_\text{cont} + \arcsin(d_\text{T} / r))$, where $\varphi_\text{cont}$ and $r$ represent the continuous azimuth angle and the range of the recently added point and $d_T$ is the clustering threshold.

\subsection{Cluster Generation}
\label{sec:cluster_generation}

Since graph $G$ contains only a few vertices and edges, it is possible to perform a new run of connected component labeling (CCL) after each arrival of a new column.
It iterates over all yet unvisited vertices $v \in V$, which then serves as the starting point for a traversal method, e.g. breadth-first search (BFS).
In each run of the traversal method, all vertices are assigned the same label as the starting vertex and marked as visited.
If during a traversal only complete vertices $v_1, v_2, ..., v_n \in V$ are found, then these vertices are removed from the linked list, which represents $V$.
In order to collect all LiDAR points belonging to this cluster the corresponding point trees $P_{v_1}, P_{v_2}, ..., P_{v_n}$ are traversed from top to bottom by using a point's $p$ child indices $p_\text{children}$ starting from the root point.

A vertex $v \in V$ or its point tree $P_v$ is considered to be complete when all points of the latest firing are distant enough from the point tree $P_v$.
More precisely, the continuous azimuth angle of the rearmost laser of the latest firing $\varphi_\text{cont, rear}$ must be far enough away from the point tree: $\varphi_\text{cont, rear} > \varphi_{\text{finished}, v}$, where $\varphi_{\text{finished}, v}$ is the completion angle of a point tree described in the previous section.

Finally, the algorithm iterates over the remaining point trees in $V$ and the minimum column index of all point roots is extracted.
This is the minimum index which has to be kept in the ring buffer.
All preceding columns are ready to be cleared as the trees can only grow to the bottom and to the right, due to the traversal sequence of cells inside the field of view (FOV), as shown in \Cref{fig:graph_of_point_trees}.

\subsection{Heuristics for Real-Time Capability}
\label{sec:heuristics}

\begin{figure}
    \centering
    \includegraphics[width=0.7\linewidth]{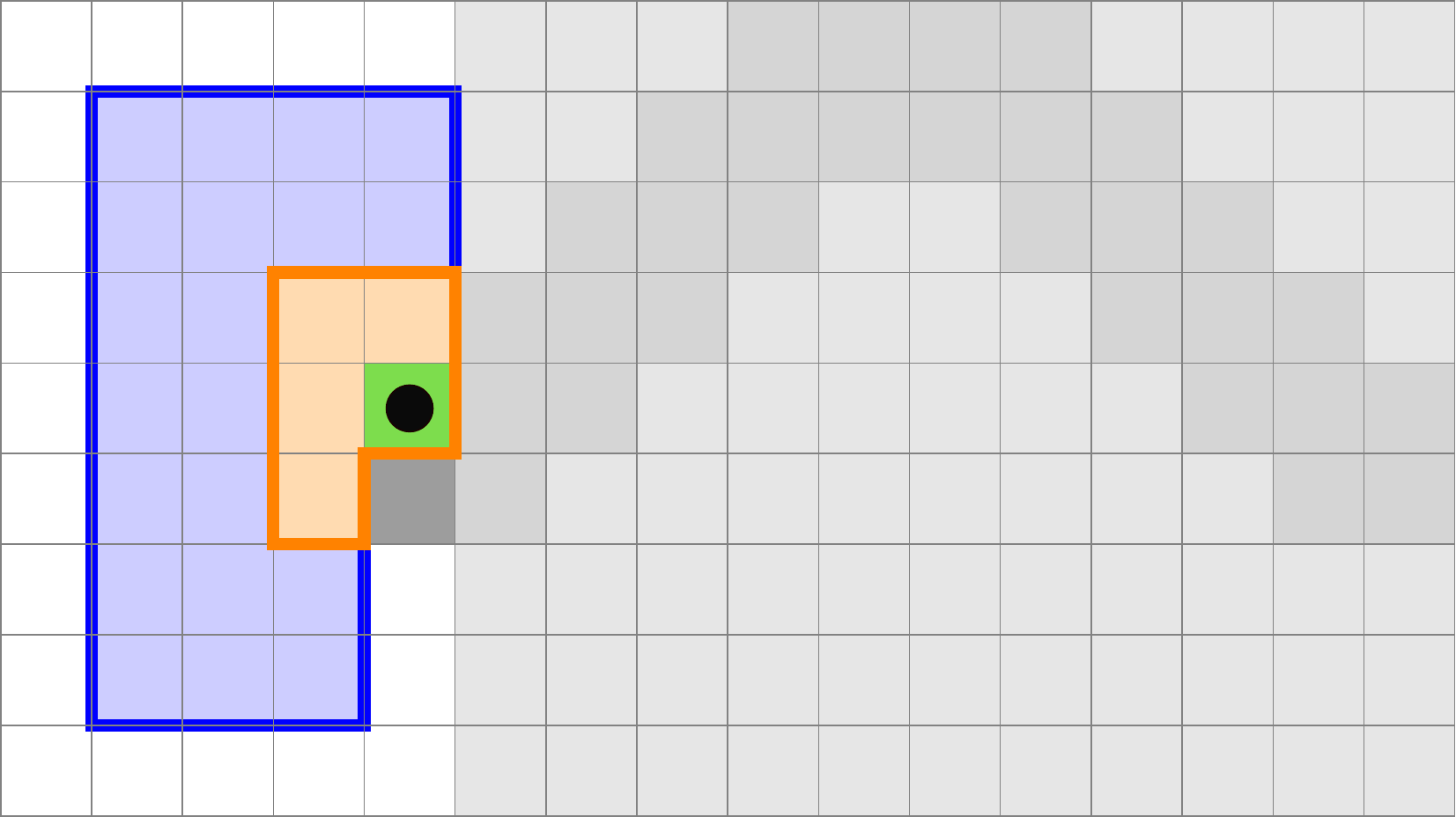}
    \caption{
    \textbf{Heuristic A:}
    Instead of evaluating the full field of view (blue), only the yellow part with configurable size is completely evaluated.
    The outer blue part is only evaluated subsequently as long no neighbor was found.
    The yellow area is important to generate a set of high-level edges $E_\text{approx}$, which is ideally as complete as $E$ in the exact solution.
    }
    \label{fig:heuristic_stop_after_association}
\end{figure}

\begin{figure}[t]
     \centering
     \begin{subfigure}[b]{\linewidth}
        \centering
        \includegraphics[trim={0 0 0 0}, clip, width=\linewidth]{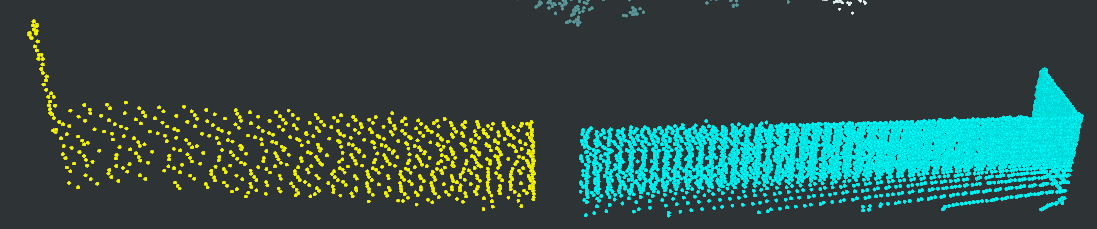}
        \caption{3D point cloud of two containers}
        \label{fig:heuristic_chessboard_a}
     \end{subfigure}
     \hfill
     \begin{subfigure}[b]{\linewidth}
         \centering
         \includegraphics[trim={0 40 0 40}, clip, width=\linewidth]{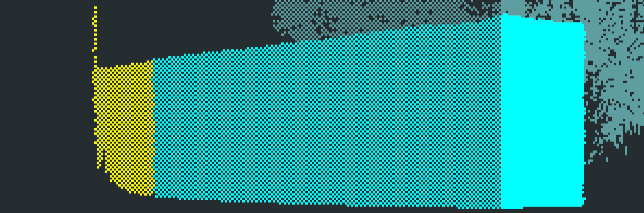}
         \caption{Same scene as crop of range image}
         \label{fig:heuristic_chessboard_b}
     \end{subfigure}
    \caption{
\textbf{Heuristic B:}
The yellow cluster and parts of the blue cluster are clustered only with cells distributed in a checkerboard pattern.
At the right side of (b) this heuristic is disabled and the range image becomes dense.
In (a), despite ignoring half of the points and the LiDAR sensor hitting the walls at a sharp angle, the points remain evenly spread without significant gaps.
Ignoring entire columns or rows in the range image would result in problematic gaps.
}
    \label{fig:heuristic_chessboard}
\end{figure}

The algorithm, as described in previous sections, delivers exact clustering results based on the Euclidean distance $d_\text{T}$.
However, the throughput, i.e. the number of points per second that can be processed, may not be enough on hardware with restricted computational resources.
This can be due to the use of embedded hardware or due to a large work load on the system.
The most computationally demanding step is iterating over the potentially large FOV for each obstacle point.
Instead of limiting the FOV size, we use a different approach, which is explained in \Cref{fig:heuristic_stop_after_association}.
The point trees $P_v$ produced with this heuristic remain equivalent to the exact method.
However, the high-level edges $E_\text{approx}$ in $G_\text{approx}$ might be incomplete, potentially affecting clustering.
However, it is observed that two point trees belonging to the same object are still very likely to be combined through another vertex.
Due to this heuristic, the FOV search can often be terminated much earlier compared to the exact method, significantly increasing the throughput.

Instead of increasing the throughput it is also possible to reduce the number of input points.
As shown in \Cref{fig:heuristic_chessboard} we achieve this by sub-sampling the range image in a checkerboard pattern.

\section{EVALUATION}
\label{sec:evaluation}

The class-agnostic nature of distance-based clustering is both an advantage and a disadvantage.
It facilitates the detection of various object types; however, without additional processing, there's no object classification available.
This becomes particularly problematic when evaluating with established object detection benchmarks \cite{caesar_nuscenes_2020, geiger_are_2012, behley_2019_semantickitti, sun_2020_scalability}, as their metrics inherently rely on object classification.
Therefore, in \cite{oh_2022_travel}, the metrics over-segmentation entropy (OSE) and under-segmentation entropy (USE) were introduced in order to evaluate over-segmentation and under-segmentation, respectively, irrespective of the class.
These metrics are evaluated using the SemanticKitti \cite{behley_2019_semantickitti} dataset.

\begin{table}
	\caption{Quantitative analysis on SemanticKitti \cite{behley_2019_semantickitti} using under- (USE) and over-segmentation entropy (OSE)}
	\label{tab:quantitative}
	\centering
	\begin{tabular}{|l|c|c|c|c|}
		\hline
		\multirow{2}{*}{\diagbox{Approach}{Metric}} & \multicolumn{2}{c|}{USE} & \multicolumn{2}{c|}{OSE}  \\
      & $\mu\downarrow$ & $\sigma\downarrow$ & $\mu\downarrow$ & $\sigma\downarrow$ \\
		\hline
		GPF + SLR \cite{zermas_2017_fast} & 117.08 & 127.40 & 301.16 & 144.29 \\
		TRAVEL \cite{oh_2022_travel} & 24.07 & 11.88 & 70.40 & 34.44 \\
     Proposed (w/ Heur. A) & \textbf{23.74} & \textbf{11.21} & 65.79 & 25.52 \\
     Proposed (w/o Heur. A) & \textbf{23.74} & \textbf{11.21} & \textbf{65.23} & \textbf{23.88} \\
		\hline
	\end{tabular}
\end{table}

In \Cref{tab:quantitative}, our approach is compared to others based on these metrics.
Furthermore, the implications of heuristic A from Section \Cref{sec:heuristics} are presented.
The distance threshold in our methods is set at $d_\text{T} = 0.7$ as it provides the minimal sum of USE and OSE scores.
It's evident that our approach, despite its low-latency nature, achieves superior segmentation.
The exact approach yields the best results.
The same approach with heuristic A, performs only slightly worse.
Compared to other approaches \cite{oh_2022_travel, zermas_2017_fast}, the issue of over-segmentation is significantly mitigated by the enlarged perceptive field of view.
However, the enlarged FOV doesn't mitigate the under-segmentation problem.
It can be observed that adding additional metrics, like considering convexity during graph edge formation, doesn't substantially reduce under-segmentation in the SemanticKitti scenarios.

In the context of our continuous clustering algorithm real-time capability refers to the average latency of all clusters.
For a cluster's reference timestamp, we utilize the timestamp of its most recent LiDAR point.
Consequently, the integration time of the cluster, which depends on the angular extent of the object, is not part of this value.
This is useful for measuring how quickly finished clusters are detected, aggregated, and published.
On our consumer grade hardware with a CPU Intel Core i7-8700K @ 3.7GHz (6 cores, 12 threads) and 32GB of RAM we achieve enough throughput to prevent the queues in the multi-threaded pipeline, as described in \Cref{sec:proposed_method}, from indefinitely growing.
In this settings the average latency of $\mu = \SI{5}{\milli\second}$ with a standard deviation of $\sigma=\SI{8}{\milli\second}$ is achieved with a Velodyne VLS-128 with 128 lasers used in an urban scenario.
However, the throughput greatly reduces when other perception algorithms are running.
Therefore we use Heuristics B to reduce the number of incoming points and Heuristic A to keep the throughput at a stable level even with other perception tasks.
While \Cref{tab:quantitative} shows that Heuristic A has only little negative impact, this can not be shown for Heuristic B, as this would greatly increase OSE as half of the points are missing compared to the ground truth instances.

\begin{figure}
     \centering
     \begin{subfigure}[b]{\linewidth}
        \centering
        \includegraphics[trim={0 40 0 60}, clip, width=\linewidth]{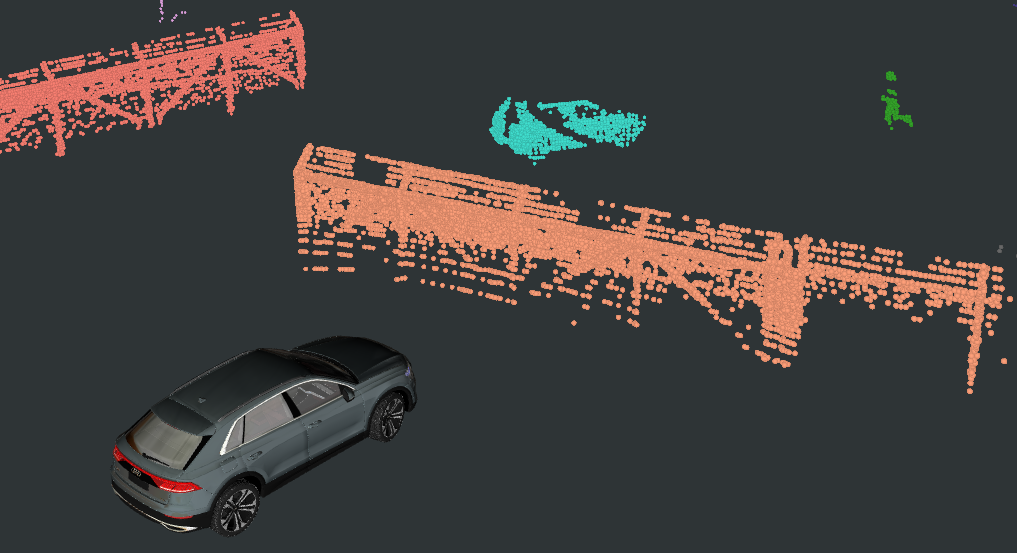}
        \caption{Challenging scene with occlusions by a fence}
        \label{fig:evaluation_fov_a}
     \end{subfigure}
     \hfill
     \begin{subfigure}[b]{\linewidth}
         \centering
         \includegraphics[trim={80 0 80 0}, clip, width=\linewidth]{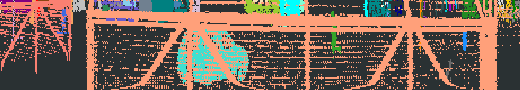}
         \caption{A 520x90 crop of the range image}
         \label{fig:evaluation_fov_b}
     \end{subfigure}
    \caption{
\textbf{Importance of FOV:} In order to correctly segment the fence (orange), the car (cyan) and the pedestrian (green) in (a) a rectangular FOV of at least 10x9 is required in (b).
}
    \label{fig:evaluation_fov}
\end{figure}

In \Cref{fig:evaluation_fov} a very challenging scenario is shown, due to heavy occlusions.
It shows the importance of being able to use a large FOV when necessary.
Some approaches use only a 4-neighborhood for each grid cell \cite{moosmann_2009_segmentation, bogoslavskyi_2016_fast}, which is very efficient.
Hierarchical approaches \cite{zermas_2017_fast, burger_2018_fast2, yang_2020_two, zhang_2022_real, oh_2022_travel} have a similar, but larger, cross-shaped FOV around the currently-evaluated cell.
Due to the aforementioned cross shape, all these approaches have difficulties with oblique and sparse structures in the range image.
Even seemingly straight obstacles, like a streetlamp, can have a horizontal jump in the depth image due to slight twists.
The approach with the largest field of view is described in \cite{zhang_2022_real}, which has a view distance of up to 3 cells in both dimensions and either direction.
However, a scenario as shown in \Cref{fig:evaluation_fov} can only be handled correctly by our approach and \cite{najdataei_2018_continuous, najdataei_2022_pi_lisco}, which are able to use a very large FOV when required.

\section{CONCLUSIONS}

In this study, we introduced an algorithm capable of clustering a continuous stream of LiDAR points in real time without prior accumulation.
This significantly reduces the latency of individual object instances. 
Simultaneously, it allows for a much larger field of view compared to existing methods, considerably enhancing the results in challenging scenarios with occlusions.

Our approach can be extended in various ways.
For instance, hierarchical divisive clustering would be interesting, with hypotheses that are hierarchically divided into sub-clusters based on decreasing distance thresholds, e.g. $d_T \in \{\SI{2}{\meter}, \SI{0.7}{\meter}, \SI{0.3}{\meter}\}$.
Thus, a subsequent perception node, like multi-object tracking, could decide which hypothesis is the most plausible one based on previous observations.

As already mentioned, it would also be very interesting to cluster not only based on Euclidean distance but also based on the similarity between the feature vectors of two points.
For this, each column would be fed into a neural network, which, with knowledge of only the current and preceding columns, could assign a feature vector to each point that may be useful for clustering.
Ideally, the neural network has spatial and temporal comprehension, allowing it to retain knowledge from the last few sensor rotations. Simultaneously, it should not be too deep, ensuring it can perform inference on columns at a sufficiently high frequency.



\bibliographystyle{IEEEtran}
\bibliography{macros/IEEEabrv,macros/additional_abrv,macros/et_al,bibliography}

\end{document}